\documentclass[runningheads]{llncs}
\usepackage{graphicx}
\usepackage{comment}
\usepackage{amsmath,amssymb} % define this before the line numbering.
\usepackage{color}

\begin{document}
% \linenumbers
\pagestyle{headings}
\mainmatter
\def\ECCVSubNumber{1435}  % Insert your submission number here

\title{Generating Memorable Images Based on Human Visual Memory Schemas} % Replace with your title

\begin{comment}
\titlerunning{ECCV-20 submission ID \ECCVSubNumber} 
\authorrunning{ECCV-20 submission ID \ECCVSubNumber} 
\author{Anonymous ECCV submission}
\institute{Paper ID \ECCVSubNumber}
\end{comment}
%******************

%\begin{comment}
\titlerunning{Generating Memorable Images}
\author{Cameron Kyle-Davidson\inst{1} \and %\orcidID{0000-0001-5144-9746} \and
Adrian G. Bors\inst{2,3}\and %\orcidID{1111-2222-3333-4444} \and
Karla K. Evans\inst{3}}%\orcidID{2222--3333-4444-5555}}
\authorrunning{C Kyle-Davidson et al.}
\institute{Dept. Of Computer Science, University of York, England \\
\email{ckd505@york.ac.uk}, \email{adrian.bors@york.ac.uk} \and
Dept. of Psychology, University of York, England \\ \email{karla.evans@york.ac.uk}}

%\end{comment}
%******************
\maketitle

\begin{abstract}
This research study proposes using Generative Adversarial Networks (GAN) that incorporate a two-dimensional measure of human memorability to generate memorable or non-memorable images of scenes. The memorability of the generated images is evaluated by modelling Visual Memory Schemas (VMS), which correspond to mental representations that human observers use to encode an image into memory. The VMS model is based upon the results of memory experiments conducted on human observers, and provides a 2D map of memorability. We impose a memorability constraint upon the latent space of a GAN by employing a VMS map prediction model as an auxiliary loss. We assess the difference in memorability between images generated to be memorable or non-memorable through an independent computational measure of memorability, and additionally assess the effect of memorability on the realness of the generated images.

\keywords{Memorability, Generative Models, Computer Vision}
\end{abstract}

\section{Introduction}

Image memorability is dependent upon both the semantic details of an image, such as its category and content, and also to a lesser degree its low level properties \cite{isola_what_2011}. Over the past few years there has been a view towards combining both psychological and computational models of memorability, such that deep learning models have now been shown to approximate human consistency for repeat-detection memory tests. However, recent human-based image memorisation studies have shown that human visual memory has both a spatial and relational component, and that memorability varies across an image. Human observers agree with high consistency on what areas of an image\cite{akagunduz_defining_2019}, or rather, compositions of objects forming structures, cause that image to be remembered, as evidenced by human memory experiments. The two-dimensional memory maps that capture this information represent a cognitive concept known as a visual memory schema. Visual memory schema maps are hypothesised to correspond to mental representations and structured knowledge human observers use to encode an image into memory, {\em e.g.} an image of a beach is memorable because it matches the cognitive schema that represents a beach in the observers' brain. Despite the high consistency of visual memory schemas among human observers, they do not appear to correlate strongly with predictions from deep models trained on repeat-detection memory test data. This indicates that the visual schema model captures additional information about memorability that is lost from other such models. Recent research work has shown that an artificial learning model can predict visual memory schema maps for scene images, \cite{kyle-davidson_predicting_2019}.

In this research study we consider generating images whose memorability feature is defined based on visual memory schemas (VMS). We propose using generative adversarial models to generate memorable or non-memorable scene images, where the structure arising in the image is based on visual memory schemas used by and extracted from human observers. Unlike any other GAN, generation of completely new images of scenes here is human data driven. The results of this study allow for the analysis of the differences that arise between memorable and non-memorable images, and provide a data driven approach to further understanding how visual schemas are structured in human cognition and memory. What is more, we additionally examine the relationship between image memorability and visual schemas on an category-by-category basis, consider the relationship between memorability and image `realness', and evaluate our results through a computational measure of memorability.

The contributions of this research study are:
\begin{itemize}
    \item A trainable generative model employing a memorability constraint based on a visual memory schema model trained upon human observer results.
    \item A further investigation examining the use of two-dimensional memorability maps as input to guide image generation.
    \item An assessment of the level of realness for the generated images, and how and why this relates to memorability defined by visual memory schemas.
    \item An evaluation of the results of our network based upon an independent measure of memorability.
\end{itemize}

\section{Related Work}

In this section we revise deep learning models such as the Generative Adversarial Network (GAN) and the concept of visual memory schemas, as well as other approaches to the modification of image memorability.

\textbf{Generative Adversarial Networks}. GANs are composed of two components: a generator which attempts to synthesise realistic data, and a discriminator which aims to separate realistic data from synthesised data \cite{goodfellow_generative_nodate}. GANs have been employed to generate photorealistic images \cite{arjovsky_wasserstein_2017,karras_progressive_2018,karras_style-based_2019} as well as for image-to-image translation tasks \cite{isola_image--image_2018,zhu_unpaired_2018}. GANs have also been employed for saliency-based scan-path prediction, indicating their ability to be used to generate and account for psychologically-based data \cite{assens_pathgan:_2018}. Certain implementations of GANs have been shown to allow meaningful and unsupervised disentanglement of high-level properties into controlled discrete variables, besides continuous data variations, such as in InfoGAN \cite{chen_infogan:_2016}. This research work employs a partially supervised method in order to control the properties related to the memorability of an image, by considering an auxiliary loss component in the optimisation function of a GAN.

\textbf{Modifying Image Memorability.} There has been a small amount of work concerning the modification of existing images to change their memorability. In the case of face images, this was accomplished through the manipulation of Active Appearance Models \cite{khosla_modifying_2013}. In more general cases, image manipulation has been achieved via memorable style transfer \cite{siarohin_how_2017} (which applies filters/enhancements to an image, such as increasing the contrast) and direct manipulation of existing face images using attention-based GANs \cite{sidorov_changing_2019}. There has also been work into extracting the `memorability' dimension from existing trained GANs, allowing for the generation of images deemed to be memorable \cite{goetschalckx_ganalyze:_2019}.
However, such an approach remains based on repeat-detection models, and requires a seed image whose memorability is then adjusted. For example, an image of a dog in a scene generated along an increasing memorability axis may lead to the image becoming entirely of the dog, with the scene information being discarded.  From a visual schema viewpoint, this may result in the application of a different schema rather than the schema applied to the original seed image, and would no longer represent scene memorability. No research work until now has investigated generating entirely new memorable scene images through the training of a deep learning network using memorability maps extracted from human observers as a constraint in their loss function, which we address in this work.

\section{Training GANs to Learn Memorable Images}

The goal of our approach is to enable a GAN to be trained in order to generate images whose memorability would be controlled by a chosen input value. To accomplish this, we first extract VMS maps from Vischema 2, and combine them with Vischema 1 to create a new, larger dataset that we use to pre-train a memorability estimation network. This network guides the GAN towards the generation of memorable or less memorable images through functioning as an auxiliary loss component based upon the average memorability score of the visual memory schema.

\subsection{Acquiring Additional Visual Memory Schemas data from Human Memorisation Experiments}

\begin{figure}[]
\centering
\includegraphics[width=0.5\textwidth]{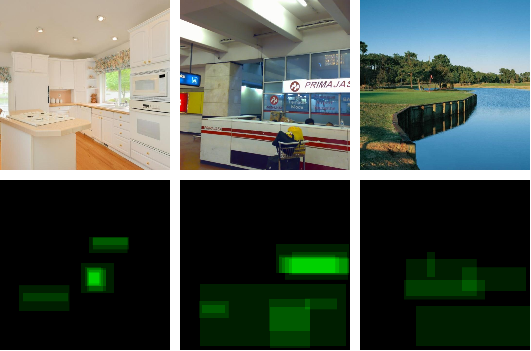}
\caption{Images used for VISCHEMA 2 dataset are shown in the top row, while in the bottom row are the image regions indicated by observers as causing them to remember the images from above.}
\label{fig:vms2}
\end{figure}

Visual Memory Schemas (VMS) \cite{akagunduz_defining_2019} are experimentally obtained two-dimensional maps that define the regions of a scene image (scene images refer to images of commonly encountered locales, \textit{e.g.} an interior image of a kitchen, a living room, or a park) which causes said image to be remembered (a true memory schema) or falsely remembered (a false schema) by human participants. These maps are highly consistent between observers, and are hypothesised to correlate with human cognitive representations of the scene. The information captured by VMS maps reveals the structures in scene images that correspond with human image memorability. Memorability as defined by VMS maps, despite their inter-observer consistency, appear to not correlate with previous, one-dimensional metrics of image memorability \cite{kyle-davidson_predicting_2019}. In the VISCHEMA 1 experiment, human observers asked to label the regions of images that caused them to remember that image, while at the same time measuring the degree to how memorable the images was to the observer, resulting in a dataset of 800 image/VMS pairs.
The VISCHEMA 2 dataset is an extension of VISCHEMA 1 containing 800 different images of the same categories as VISCHEMA 1 but lacking the VMS information \cite{kyle-davidson_predicting_2019}. 800 VMS maps are gathered for the VISCHEMA 2 dataset, through human observer experiments, thus creating a  1600 image/VMS pair dataset. We call this dataset VISCHEMA PLUS, and use it to train the auxiliary loss function in our generative approach. We describe this auxiliary function in the next section.

\begin{figure}[]
\vspace*{-15pt}
\centering
\includegraphics[width=0.5\textwidth]{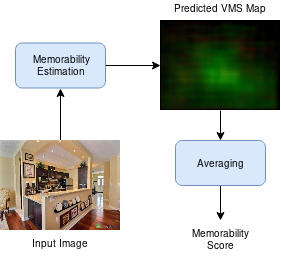}
\caption{Calculating memorability scores by using VMS map prediction.}
\label{fig:vae}
\end{figure}

\vspace*{-13pt}
\subsection{Memorability Model}

The assessment of image memorability is performed by using a Visual Memory Schema prediction model developed in \cite{kyle-davidson_predicting_2019}, which is based on the Variational Autoencoder (VAE). Following training, given an image, the VAE is used to predict its corresponding VMS. We implement this model, and train it on the VISCHEMA PLUS dataset (1600 image/VMS pairs). The output of this model is a two-dimensional VMS map, which is reduced to a single score that indicates the `memorability' of any given input image, as illustrated in Fig.~\ref{fig:vae}. This VMS map represents a predicted combination of multiple human observations for that image. We only consider the `memorability' channel of the VMS maps (true schemas), and do not make use of the `false memorability' (false schemas) information. Its training relies on the following VAE' loss function~:
\begin{equation}
L(\theta,\phi) = -E_{{\bf z} \sim q_{\theta}({\bf z}|{\bf x})}[\log p_{\phi}({\bf x}|{\bf z})] + KL(q_{\theta}({\bf z}|{\bf x})||p({\bf z})),
\label{eq1}
\end{equation}
where the former term represents the log-likelihood of VMS reconstruction by using the decoder network and the latter represents the Kullback-Leibler (KL) divergence between the variational distribution $q_{\theta}({\bf z}|{\bf x})$ and the prior $p({\bf z})$. Here ${\bf x}$ represents the VMS data, while ${\bf z}$ are the latent variables inferred by the encoder, and where $\theta$ and $\phi$ represent the parameters of the VAE's encoder and decoder networks, respectively.

\begin{figure*}
\vspace*{-10pt}
\begin{center}
\includegraphics[width=0.95\textwidth]{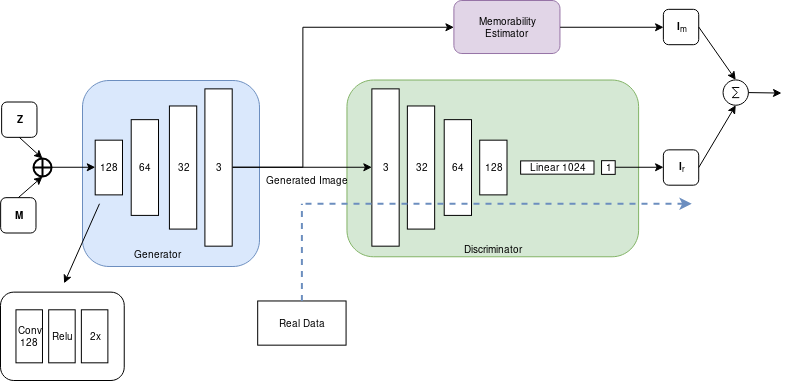}
\end{center}
\vspace*{-10pt}
   \caption{Memorability-constrained image generation model architecture. PixelNorm and Minibatch Standard Deviation layers omitted for clarity.}
\label{fig:memgen}
\end{figure*}
\vspace*{-10pt}
\subsection{Memorable Image Generation System}

In the following we adapt an improved Wasserstein GAN (WGAN) \cite{gulrajani2017improved} model for generating memorable images. The image generation network $G$, corresponding to the generator from WGAN, aims to synthesise an image $\mathbf{\hat{I}}$ using as inputs random variables $\mathbf{Z}$, which defines the latent space of the GAN, while $\mathbf{M}$ acts as our memorability constraint:
\begin{equation}
\mathbf{\hat{I}} = G(\mathbf{Z}, \mathbf{M}).
\label{eq2}
\end{equation}
The output of the generator is a generated image $\mathbf{\hat{I}}$, whose memorability score is as close to $\mathbf{M}$ as possible. Both $\mathbf{Z}$ and $\mathbf{M}$ are drawn from Gaussian distribution. The generator is constrained by both the discriminator $D$ and by an auxiliary memorability function ${\cal M}$, defining $\mathbf{\hat{I}}_m = {\cal M}(\mathbf{I})$ where $\mathbf{\hat{I}}_m$ is an estimation of the images memorability. We characterise ${\cal M}$ through a neural network. The proposed learning model for generating memorable images consists of a Generator, a Discriminator, and a Memorability evaluation network and its diagram is shown in Fig.~\ref{fig:memgen}. While the generator creates memorable images, the discriminator evaluates the `realness' of the generated images, and the auxiliary memorability network evaluates whether the memorability of the generated image matches the memorability defined by $\mathbf{M}$. The input consists of two latent variables $\mathbf{Z}$ and $\mathbf{M}$ which are concatenated before being passed to the generator. The discriminator $D$ used is that of the improved Wasserstein GAN \cite{gulrajani2017improved} which employs a penalty term on the critic loss yielding better performance and stability when compared to the classical GAN.

\subsection{VMS-Based Loss Function}

In the following we develop a model which would enforce the characteristics of being memorable in generated images. In order to do so we modify the loss function of the WGAN, by adding an additional component which calculates the loss between the desired and generated memorability for a given image. The loss function is defined as~:
\begin{equation}
    L = L_{WGAN} + \alpha 
    \left[\frac{1}{N}\sum_{i=1}^{N}(\mathbf{M}_i - \mathbf{\hat{I}}_m)^{2}\right]
    \label{eq5}
\end{equation}
where $N$ is the batch size, $\alpha$ represents the contribution of the memorability loss and  $L_{WGAN}$ is the standard Improved Wasserstein loss \cite{gulrajani2017improved} given by~:
\begin{equation}
   L_{WGAN} = 
     \underset{\hat{\mathbf{I}} \sim \mathbb{P}_g}{\mathbb{E}} [D(\hat{\mathbf{I}})] - \underset{\mathbf{I} \sim \mathbb{P}_r}{\mathbb{E}}[D(\mathbf{I})] + \lambda \underset{\tilde{\mathbf{I}} \sim \mathbb{P}_{\tilde{\mathbf{I}}}}{\mathbb{E}}[(||\nabla_{\tilde{\mathbf{I}}}D(\tilde{\mathbf{I}})||_{2} - 1)^{2}]
    \label{eq:wass}
\end{equation}
where $\mathbb{P}_g$, represents the probability of the generated data and $\mathbb{P}_r$ is the probability of the real data and $\mathbb{P}_{\tilde{I}}$ refers to a sampling distribution used to calculate the gradients inside the discriminator. The additional term, defining the memorability for the generated images and controlled by the $\lambda$ hyperparameter prevents the gradients inside the discriminator from becoming non-Lipschitz continuous whereas the previous two terms evaluate the Earth-Mover distance between the generated and real distributions. The Earth-Mover distance is minimised through the same optimization procedure used for training the WGAN model. This has the effect of matching two complex distributions through simulating the movement of the earth from heaps into corresponding holes through an optimal transfer procedure. We alter the loss function distance through an additional term, constraining the image generation by both `realness' and memorability, simultaneously. The additional loss term from equation (\ref{eq5}) enforces that the memorability of the images generated, modelled by their estimated VMS, corresponds to the real VMS for the same images.

\subsection{The training procedure}

During the training, $\mathbf{Z}$ and $\mathbf{M}$ are sampled randomly from Gaussian distributions and passed to the generator $G$. When training the discriminator $D$, $\mathbf{M}$ is discarded, as it is only necessary for training the generator. For training the generator, $\mathbf{M}$ is used to calculate the final term of the loss function from equation (\ref{eq5}). This has the effect of penalising the generator if the generated images are not of a similar memorability to that chosen randomly as defined by $\mathbf{M}$. For example, if the image was intended to be memorable while actually it is not memorable, the generator loss will increase.

\vspace*{-0.3cm}
\section{Experimental results}
\vspace*{-0.1cm}

Our model is implemented and trained under Keras and Tensorflow, using the Adam optimizer \cite{kingma2014adam}, considering the the learning rate $\lambda = 0.0001$, and other hyperparameters as $\beta_1 = 0.5$, $\beta_2 = 0.99$, and $\epsilon = 10^{-8}$. The batch size is set at 64, and the model is trained for 320 epochs. We apply pixel normalisation after every convolution layer in the generator to prevent excessive signal magnitudes, and make use of a minibatch standard deviation layer after the first convolution layer in the discriminator in order to enhance variation. The decision to implement these techniques is inspired by the results from Karras {\em et al.} from \cite{karras_progressive_2018}. The primary dataset trained on is LSUN-Kitchen \cite{yu2015lsun}, of which we use a subset of 120,000 images reduced to a resolution of $128^{2}$ (we also show some results from LSUN-living-room and LSUN-cathedral). 

\subsection{Validity of using Visual Memory Schemas for predicting image memorability}

Visual memory schema maps capture spatial and relational components of memory, and hence contain additional information compared to single-score based image memorability methods. VMS maps have been shown to not correlate strongly with other, more basic memorability prediction methods, nor does saliency completely explain visual memory schemas \cite{kyle-davidson_predicting_2019}. 

\begin{table}[]
\centering
\begin{tabular}{lcc}
\textbf{}                                                       & \multicolumn{2}{c}{\textbf{Consistency}}                        \\
\textbf{Category}                                               & \multicolumn{1}{l}{VISCHEMA 1} & \multicolumn{1}{l}{VISCHEMA 2} \\
Isolated                                                        & 0.556                          & 0.447                          \\
Populated                                                       & 0.624                          & 0.562                          \\
\begin{tabular}[c]{@{}l@{}}Public \\ Entertainment\end{tabular} & 0.706                          & 0.661                          \\
Work/Home                                                       & 0.674                          & 0.57                           \\
Kitchen                                                         & 0.628                          & 0.479                          \\
Living Room                                                     & 0.568                          & 0.446                          \\
Small                                                           & 0.611                          & 0.525                          \\
Big                                                             & 0.637                          & 0.595                    
\end{tabular}
\vspace{0.5cm}
\caption{Vischema 1 and Vischema 2 consistency, per category. Certain categories are more consistent than others.}
\label{tab:visccons}
\end{table}

The VISCHEMA 1 and 2 datasets contain a variety of images, grouped in the following categories~: Isolated, Populated, Public, Entertainment, Work/Home, Kitchen, Living Room, Small and Big.
In the following, we examine the consistency of both VISCHEMA 1 and VISCHEMA 2 on a category-by-category basis, shown in Table \ref{tab:visccons}. Consistency is calculated by taking 25 splits of the data (one split creating two VMS maps for each image, each built from an equal division of human annotation data) and correlating the resulting VMS maps against each other. In all cases the correlation is positive, and in many cases, strongly positive. As it can be observed from this table, the most consistent category is `Entertainment,'' which contains images of fairgrounds and playgrounds.

In order to validate that visual memory schemas can capture image memorability we calculate the signal strength of observers' memory for the images by using the sensitivity index, also called the $D'$ measure:
\begin{equation}
    D' = Z(HR) - Z(FAR),
    \label{eq:dp}
\end{equation}
where $Z(\cdot)$ is the inverse of the cumulative distribution function of the standard Gaussian, $HR$ is the hit rate and $FAR$ is the false alarm rate.  The sensitivity index, $D'$ is a measure from signal detection theory that represents the strength of a given signal, in our case characterising the memorability of the image to the human observers. The results for the $D'$ scores are provided in Table~\ref{tab:viscdprime} and these results relate to the consistency of the visual schemas. It can be observed from this table that certain categories display stronger consistency signals than others indicating that such categories are inherently more memorable in the tested humans, during the image memorization experiments.

\begin{table}[]
\centering
\begin{tabular}{lcc}
\textbf{}                                                       & \multicolumn{2}{c}{\textbf{D Prime}}                            \\
\textbf{Category}                                               & \multicolumn{1}{l}{VISCHEMA 1} & \multicolumn{1}{l}{VISCHEMA 2} \\
Isolated                                                        & 1.008                          & 0.692                          \\
Populated                                                       & 1.47                           & 1.197                          \\
\begin{tabular}[c]{@{}l@{}}Public \\ Entertainment\end{tabular} & 2.037                          & 1.813                          \\
Work/Home                                                       & 1.896                          & 1.38                           \\
Kitchen                                                         & 1.602                          & 1.257                          \\
Living Room                                                     & 1.725                          & 1.252                          \\
Small                                                           & 1.52                           & 1.4                            \\
Big                                                             & 1.741                          & 1.7                           
\end{tabular}
\vspace{0.5cm}
\caption{D-Prime Analysis of of human memory for each category in the Vischema 1 and Vischema 2 Datasets. High values clearly indicate that the memory signal for the given image category is strong and thus image memorability is high. Certain categories have stronger signals than others, possibly due to easier or more available encoding schemas among the human participants.}
\label{tab:viscdprime}
\end{table}

\begin{figure}[]
\centering
\includegraphics[width=0.6\textwidth]{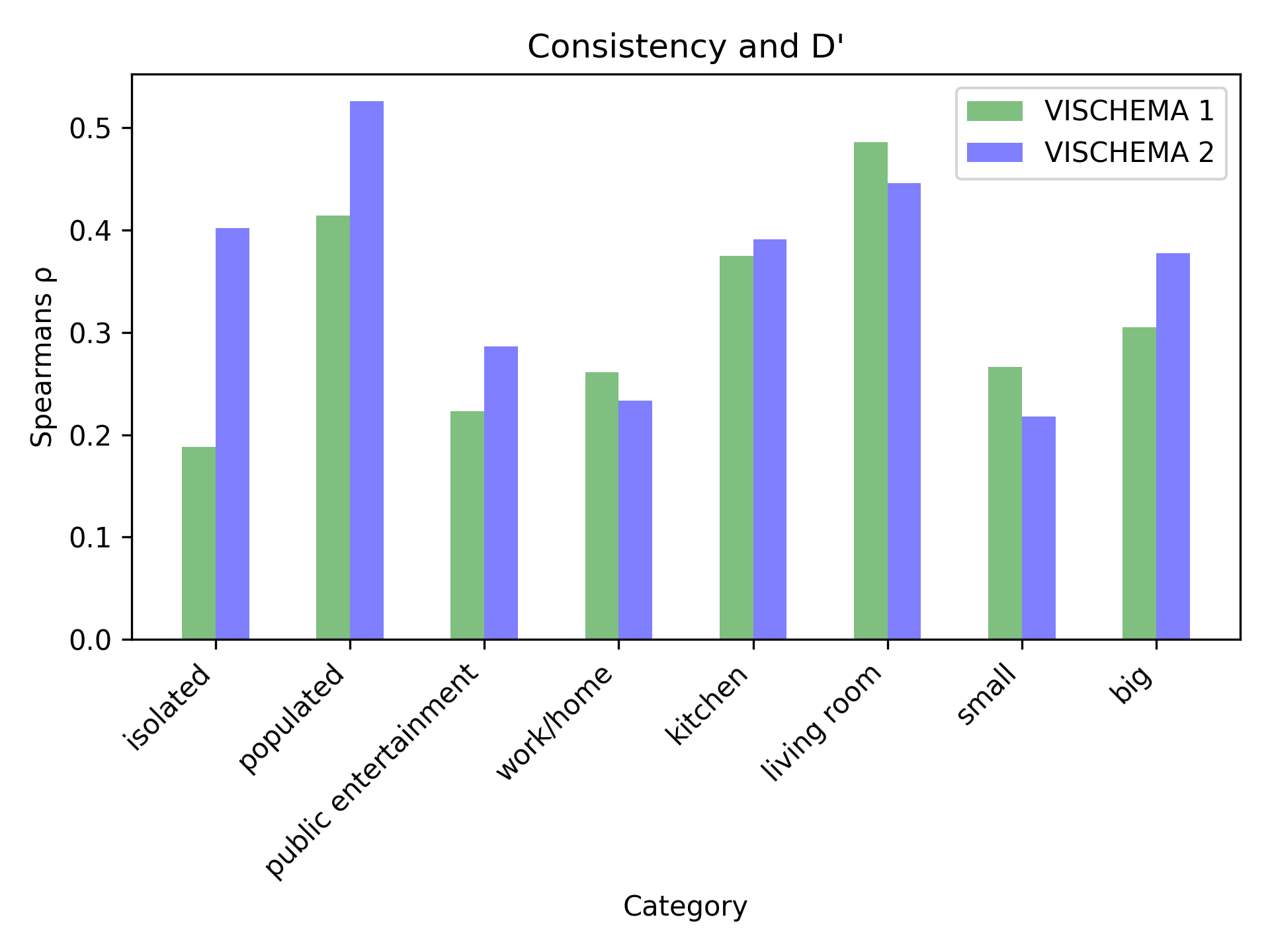}
\caption{A plot of consistency against memorability signal strength. Similar correlations between datasets indicates the reliability of using Visual Memory Schemas.}
\label{fig:dprime_vs_consist}
\end{figure}

When comparing the $D'$ results with that of consistency, we find that overall there is a strong positive correlation between the image category memorability and the category consistency for VMS maps, of $\rho = 0.83$, p \textless ~0.05, and 0.76, p \textless ~0.05, for VISCHEMA 1 and 2, respectively. When comparing this correlation for each datapoint rather than each category, we see a weaker, yet still positive correlation, shown in Fig~\ref{fig:dprime_vs_consist}. The overall high VMS consistency and it's positive correlation with the image memorability signal (measured by $D'$) indicates that VMS maps are a good descriptor of memorability and hence useful as an evaluation metric for our approach. We also evaluate the relationship between the average of the VMS map memorability channel with the image memorability signal and find a weaker, yet positive and significant relationship of $\rho = 0.307$  and 0.35, $p < 0.05$, for VISCHEMA 1 and 2, respectively. This indicates that the average VMS maps retain information about the overall memorability of the image, and are suitable for use in our generative network by enforcing VMS-based memorability.

\subsection{Results when modulating image memorability}

\begin{figure}[]
\centering
\includegraphics[width=0.49\textwidth]{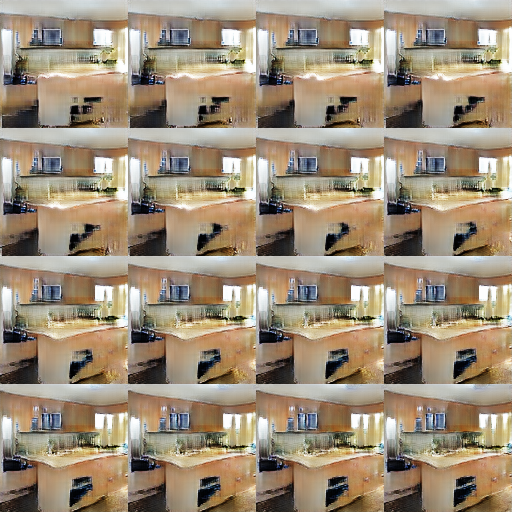}
\includegraphics[width=0.49\textwidth]{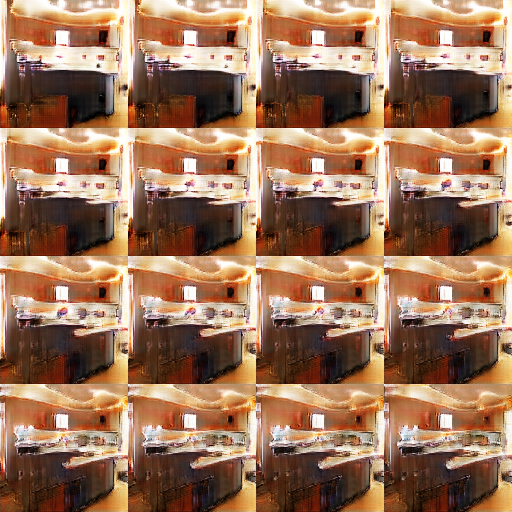}
\caption{Generated images when fixing $\mathbf{Z}$, where the sequence of generated images is displayed from left to right, top to bottom, while memorability $\mathbf{M}$ is varied from low to high. We can observe a reduction in the semantic instability (unclear structure) as memorability increases. It can be observed that as memorability increases, the `kitchen' schema emerges, where the generated images increasingly resemble the structure of a typical kitchen.}
\label{fig:ex1}
\end{figure}

\begin{figure}[]
\centering
\includegraphics[width=0.3\textwidth]{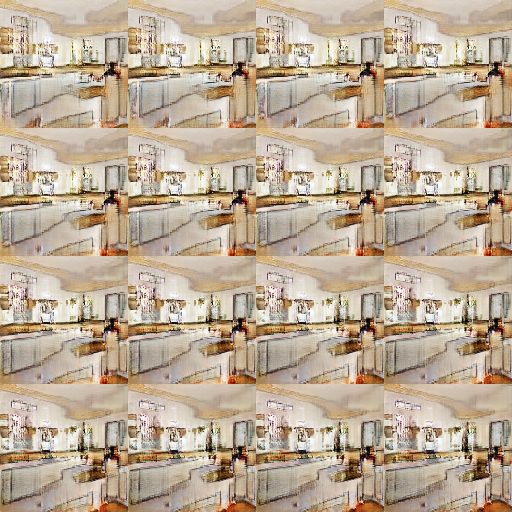}
\includegraphics[width=0.3\textwidth]{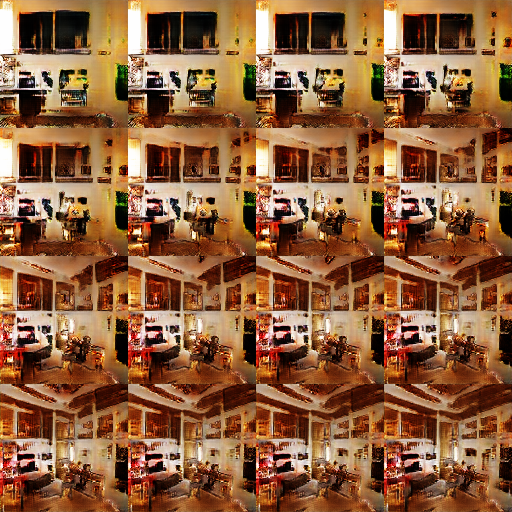}
\includegraphics[width=0.3\textwidth]{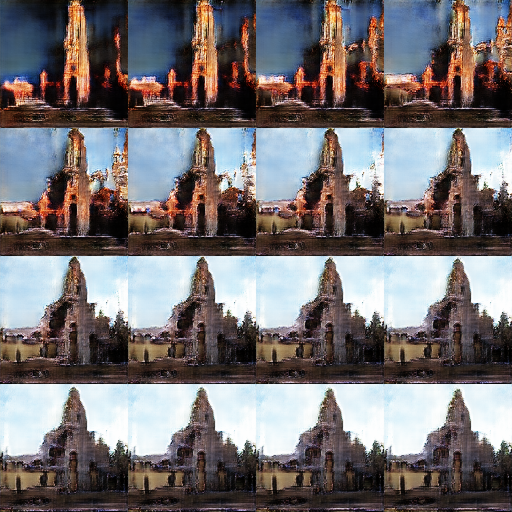}

\caption{Additional generative results with fixed $\mathbf{Z}$ and increasing $\mathbf{M}$ for kitchens, living rooms, and cathedrals. Results appear to generalise across different image sets. More memorable images bring with them increased structure, clarity, and detail; and hence better match human cognitive schemas.}
\label{fig:ex2}
\end{figure}

We generate a range of images characterised by various levels of memorability, from low to high, by fixing $\mathbf{Z}$ and varying $\mathbf{M}$. We plot these images in ascending memorability in order to examine the variation between non-memorable and memorable images.  Figures \ref{fig:ex1} and \ref{fig:ex2} show the  generation of two images obtained by fixing $\mathbf{Z}$ and varying $\mathbf{M}$ from low memorability to high memorability, with the images displayed in sequential order from top-left to bottom-right. These results explore the generated image memorability space, showing a smooth exploration of the manifold. It can be observed from these images that clear differences emerge between images when increasing the memorability constraint. In both cases, as memorability increases, semantic details and a more `kitchen-like' appearance emerges. The low memorability cases appear to display semantic `noise' representing a collection of mismatched features with loose spatial relations. The less memorable image may display the typical elements of a kitchen, but lacks structure, or rather the correct spatial relationship between the elements. It appears that by defining visual memory schemas as constraints of memorability results not only in the appearance of memorable semantic details, but also enforces spatial relationships between these details. This lends evidence that VMS maps capture semantic details and structures which match learned schemas held in human cognition.

\subsection{The realism of generated images when modulating memorability}
\begin{figure}[h]
\centering
\includegraphics[width=0.6\textwidth]{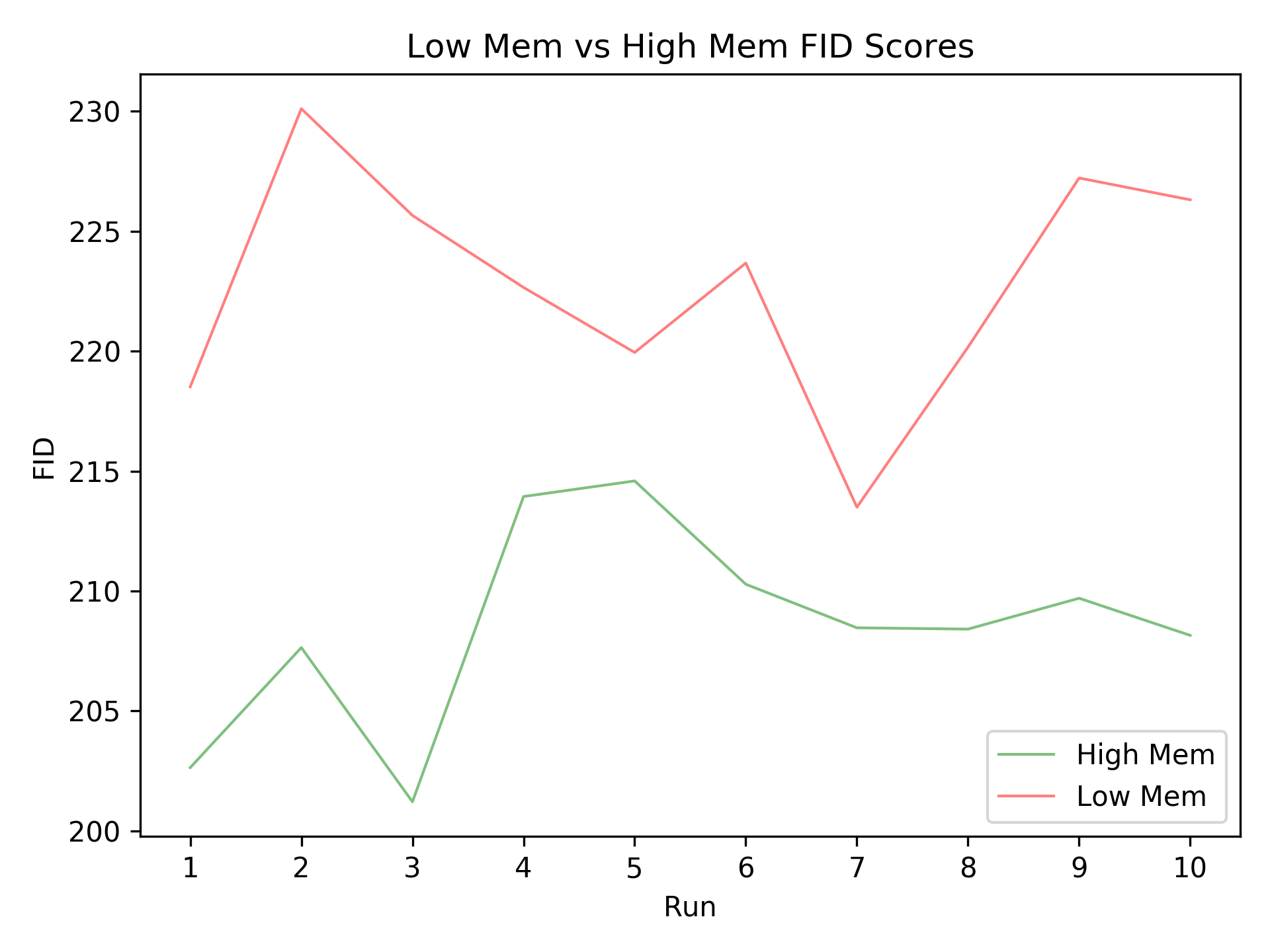}
\caption{FID scores for 10 different sets of 100 low memorability and 100 high memorability generated images. Lower values indicate more `realistic' images.}
\label{fig:fids}
\end{figure}

We generate 10 sets of 100 low-memorability images and 100 high-memorability images and calculate the Freschet Inception Distance (FID) \cite{heusel2017gans} between the generated images and real images from the training dataset. The FID score is a frequently used measure of how close a generated image is to the training set, implicitly showing how `real' is an image. We find that the images generated to be less memorable are less realistic as measured by the FID, when compared to images generated to be memorable, as shown in the plot from Fig.~\ref{fig:fids}. This difference is due to the fact that more memorable images contain additional semantic details and more structure. It should be noted that `realism' is not necessarily a prerequisite for memorability. A person familiar with abstract art is likely to have established `abstract art schemas' that enhance their ability to memorise the the art being observed, without consideration for how `realistic' that art is. Thus it is not the realism of the generated images, but how closely they fit learned schemas, that defines memorability.

\subsection{Independent memorability evaluation of generated images}

\begin{figure}
\begin{tabular}{cc}
\includegraphics[width=0.49\textwidth]{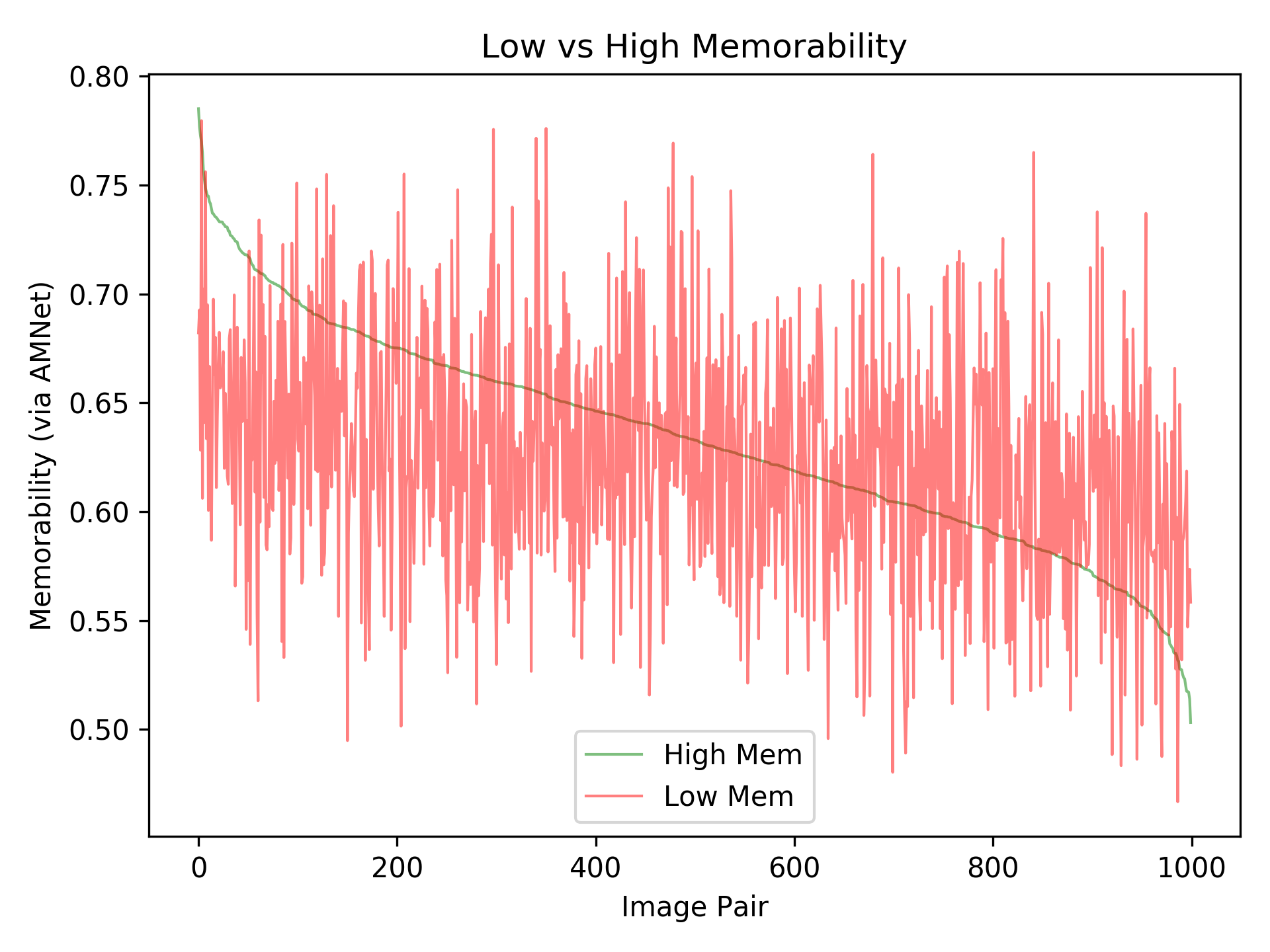} &
\includegraphics[width=0.49\textwidth]{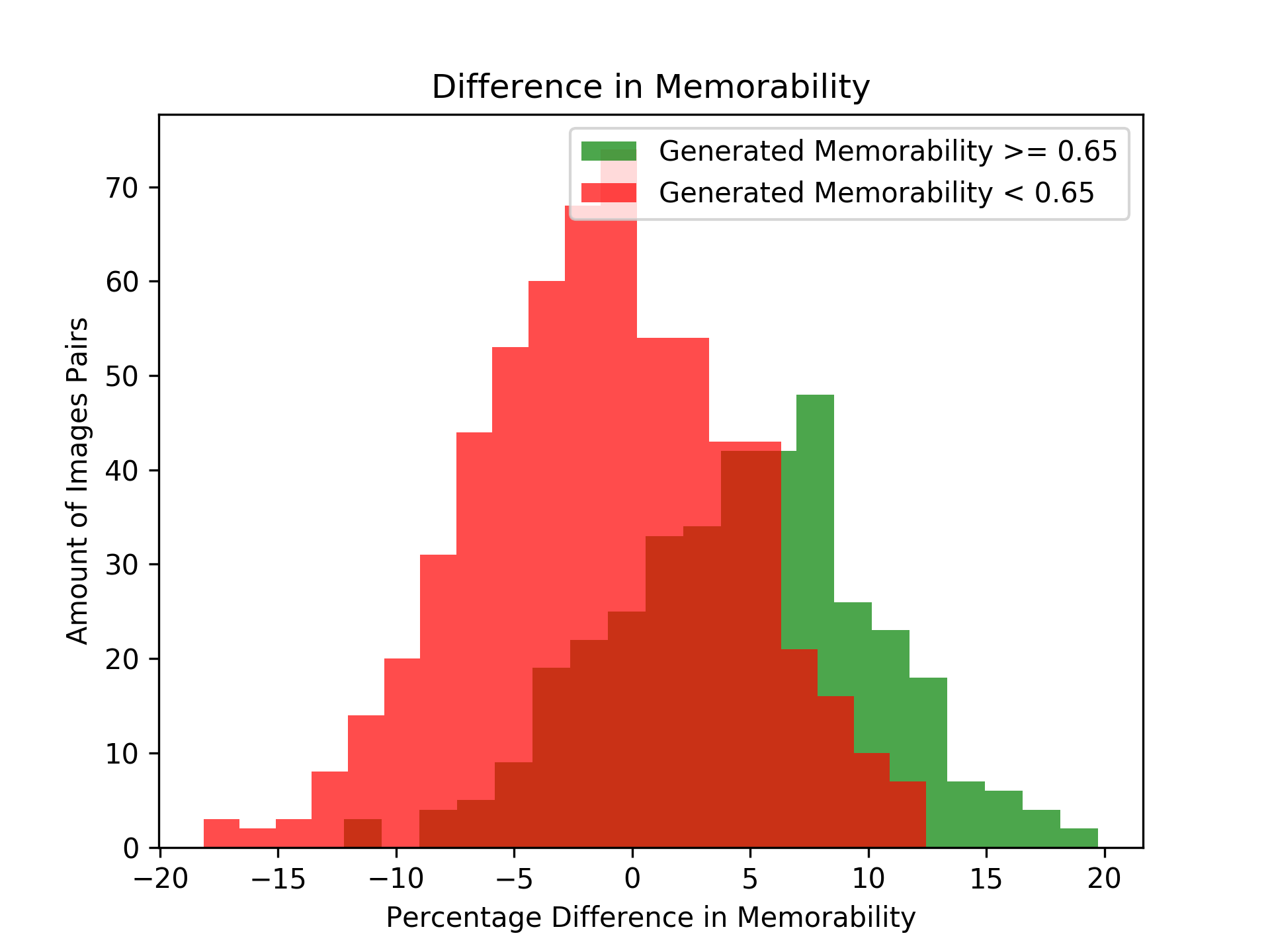} \\
(a) Prediction results &  (b) Percentage differences in \\ 
& memorability between image pairs.
\end{tabular}
\caption{AMNet prediction results for 2,000 images grouped in 1,000 pairs, generated by considering the same $\mathbf{Z}$ code but with low and high memorability, respectively.}
\label{fig:amnet_curve}
\end{figure}

In the following we generate 2,000 images by considering a fixed $\mathbf{Z}$ in 1,000, and setting $\mathbf{M}$ either very low or very high in order to create sets of pairs of images where only memorability varies while the $\mathbf{Z}$ code does not. We then evaluate these images using AMNet \cite{fajtl2018amnet}, an independent and state-of-the-art memorability prediction network. AMNet predicts the memorability of images on a scale between 0 and 1.0, allowing us to calculate the difference between our intended memorable and less memorable images. The results are shown in Fig.~\ref{fig:amnet_curve}. In many cases, the image generated to be more memorable is predicted as being more memorable than the non-memorable images generated. As overall image memorability decreases, the efficacy of our method decreases, which we show in Fig. \ref{fig:amnet_curve}a. This indicates it is more difficult to influence the memorability, or generate, certain scene images when that image is already not particularly memorable. As it can be observed from \ref{fig:amnet_curve}b, we find when the image generated to be memorable has a predicted memorability above 0.65, it can be observed that 79.5\% of the pairs have a positive difference in memorability. When memorability falls below 0.65, only 40.7\% of the pairs have a positive difference in memorability, where a `positive difference in memorability' indicates that the image generated to be memorable is predicted as more memorable than the image generated to be non-memorable. This indicates that our network has successfully learned to modulate memorability based upon Visual Memory Schemas such that the effect is verifiable by an independent memorability model, especially in cases where the generated image is highly memorable.

\subsection{Spatial Memorability with Targeted Visual Memory Schemas}

\begin{figure}[h]
\centering
\includegraphics[width=0.8\textwidth]{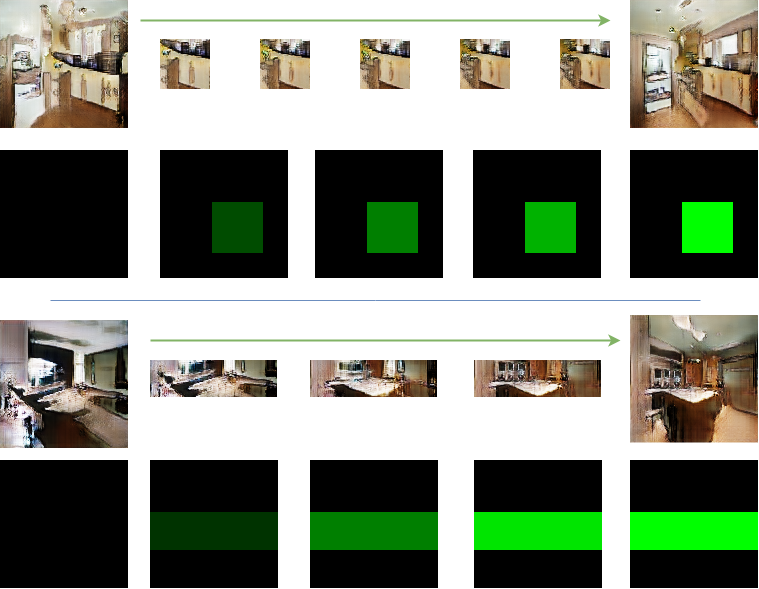}
\caption{Examples generated from a spatially-aware VMS memorability network. The latter half of each example shows the intensity (memorability) and location of the targeted region. The former half displays the image with a low memorability on the left, progressing to the high memorability image on the right. The images in between reveal semantic changes occurring within the VMS-masked region of the image.}
\label{fig:spatial}
\end{figure}

A single averaged score does encode VMS data in such a way that the relationship with the memorability signal extracted from the human observers is maintained. However, this does not capture any spatial information. As visual memory schema maps reveal, not all regions of the image are equally memorable and in many cases memorability is concentrated on certain structures inside the image, which we hypothesise to carry semantic information that matches corresponding cognitive structures (schemas) in the observers brain. With this in mind, we modify our generative model to take as input a $10 \times 10$ pixel map describing the intended spatial memorability of the generated image, and for our auxiliary loss, instead of averaging the predicted VMS map, we resize it to fit $10\times10$ pixels. We keep all other parameters of our network while using the same loss function from  eq. (\ref{eq5}). Fig. \ref{fig:spatial} shows some examples from this approach.

\begin{figure}
\begin{tabular}{cc}
\includegraphics[width=0.49\textwidth]{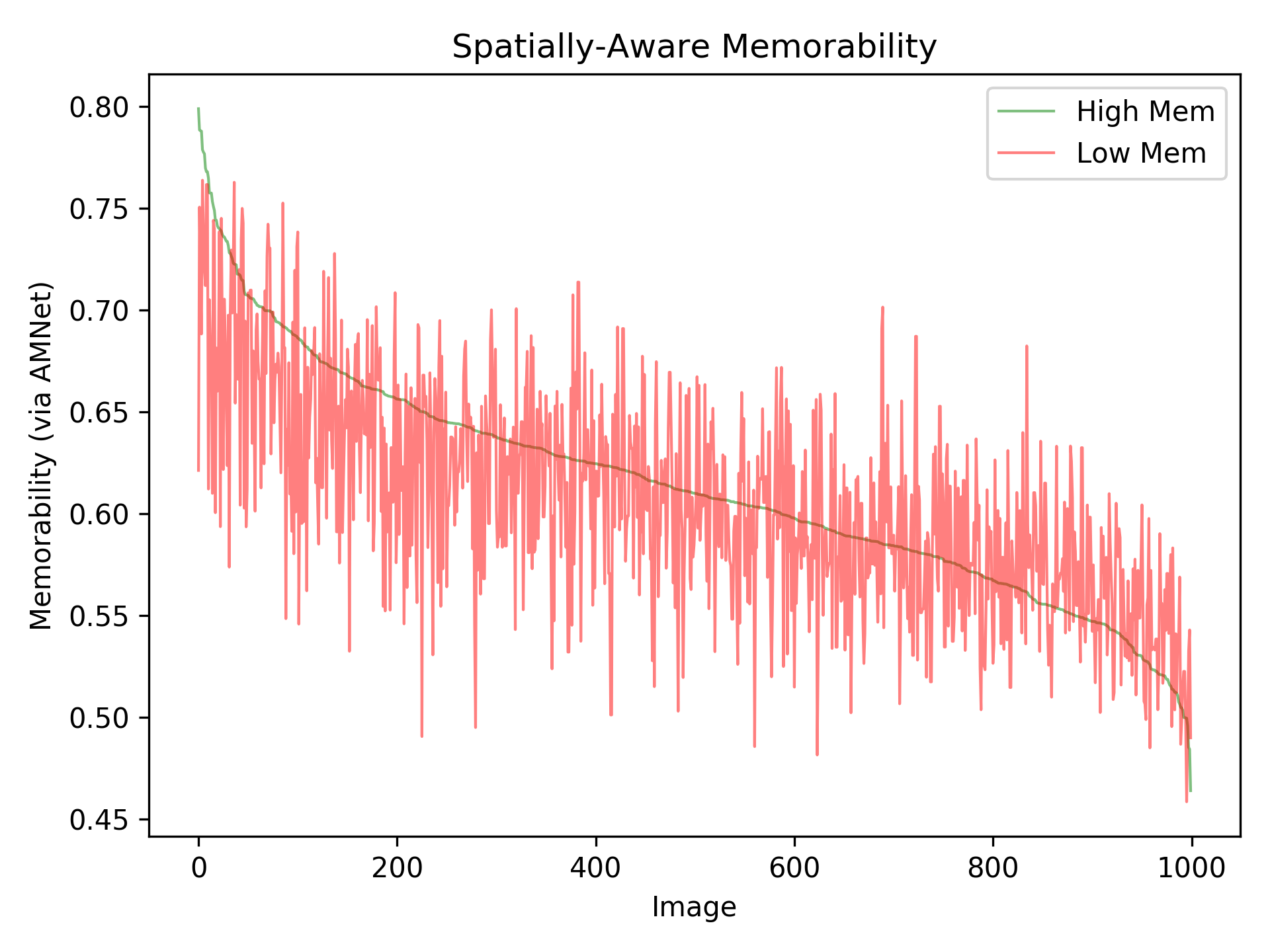} &
\includegraphics[width=0.49\textwidth]{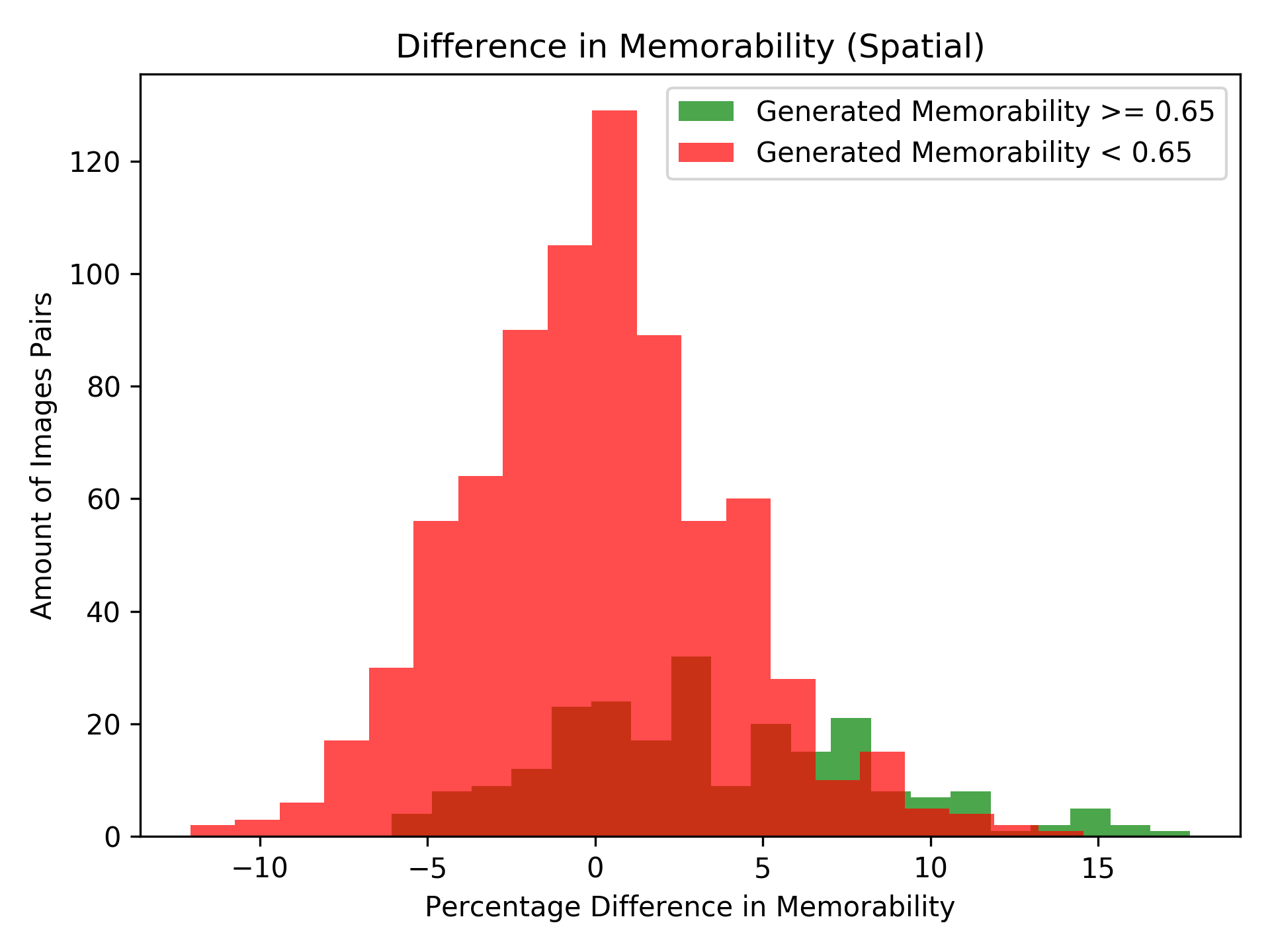}  \\
(a) Results for spatially aware memorability & (b) Histograms of differences
\end{tabular}
\caption{Comparing high/low memorability pairs of images. We find that generating images using spatial memorability data leads to generated images that display greater differences in memorability than those generated by a single score.}

\label{fig:spatial_amnet_compare}
\end{figure}

When modifying the memorability of an image through an artificial target VMS, representing `input schema', it is not just the region targeted within the map that changes. As visual memory schemas capture global information about the structures in an image, and are dependent upon them, this is not unexpected. As with the single-value score, we see clearer semantic structure arising with more memorable input schemas, and empirically we see more alteration of structure within the targeted region than outside it. We find that by using spatial memorability data to generate images, results in more robust differences between less memorable and more memorable images. 74.5\% of image pairs where the `highly memorable' image has an AMNet predicted memorability above 0.65 show a positive difference in memorability. Where the highly memorable image falls below 0.65 memorability, 50\% of pairs show a positive difference in memorability, representing a 10\% increase compared to the single-value approach. This is not meant to be a comprehensive study on generating images using spatial memorability information, but the results from Fig. \ref{fig:spatial_amnet_compare}, indicate such an approach generalises better than using a single score, even when that score is based on a visual memory schema. This is, to our knowledge, the first research study exploring the use of human spatial memory data to generate memorable images. 

\section{Conclusions}
\label{sec:conc}

 This approach is the first example of a GAN specifically trained from scratch to generate memorable scene images that employs two-dimensional memorability data gathered from human experiments. A powerful generative model such as a Wasserstein GAN (WGAN) is adapted for generating memorable images.The  loss function for WGAN was constrained by a memorability metric constraint, depending on Visual Memories Schemas (VMS) associated to the image. We investigate the relationship between the VMS map consistency and image memorability and show that VMS maps are a valid choice for a memorability metric. We evaluate the results of our model and inspect the effects of high versus low memorability on the `realness' of the generated images. Moreover, by using an independent memorability prediction network we find that the images generated to be memorable tend to be predicted as more memorable than the images generated to be less memorable. We additionally investigate the use of entire spatial memorability maps for the generation of memorable images and find such an approach to be more robust than single value memorability alone. This provides evidence that we can manipulate the memorability of generated images in a meaningful way. Image memorability defined based on visual memory schemas appears to control both the emergence of semantic details and the spatial relationships created between these details. These results help explain what Visual Memory Schema maps capture, how semantic structure contributes to the visual memory signal in human cognition, and consequently, image memorability.

%\clearpage
\bibliographystyle{splncs04}
\bibliography{generating_memorable}
\end{document}